\documentclass[journal,comsoc]{IEEEtran}
\usepackage[T1]{fontenc}
\usepackage{cite}

\ifCLASSINFOpdf
   \usepackage[pdftex]{graphicx}
\else
   \usepackage[dvips]{graphicx}
\fi

\usepackage{amsmath}
\usepackage[cmintegrals]{newtxmath}

\usepackage[numbers]{natbib}
\usepackage{booktabs}
\usepackage{multirow}
\usepackage{multicol}
\usepackage{amsmath}
\usepackage{algpseudocode}
\usepackage{mathtools}
\usepackage[linesnumbered,ruled,vlined]{algorithm2e}

\mathchardef\mhyphen="2D

\begin{document}
%
\title{An LLM-Driven Closed-Loop Autonomous Learning Framework for Robots Facing Uncovered Tasks in Open Environments}

\author{Hong~Su
\IEEEcompsocitemizethanks{\IEEEcompsocthanksitem H. Su is with the School of Computer Science, Chengdu University of Information Technology, Chengdu, China.\\
 E-mail: suguest@126.com. \\
\protect\\
}
\thanks{}}

\markboth{Journal of \LaTeX\ Class Files,~Vol.~14, No.~8, August~2015}%
{Shell \MakeLowercase{\textit{et al.}}: Bare Demo of IEEEtran.cls for IEEE Communications Society Journals}
%

\maketitle

\begin{abstract}
Autonomous robots operating in open environments need the ability to continuously handle tasks that are not covered by predefined local methods.
However, existing approaches often rely on repeated large-language-model (LLM) interaction for uncovered tasks, and even successful executions or observed successful external behaviors are not always autonomously transformed into reusable local knowledge.
In this paper, we propose an LLM-driven closed-loop autonomous learning framework for robots facing uncovered tasks in open environments. The proposed framework first retrieves the local method library to determine whether a reusable solution already exists for the current task or observed event. If no suitable method is found, it triggers an autonomous learning process in which the LLM serves as a high-level reasoning component for task analysis, candidate model selection, data collection planning, and execution or observation strategy organization. The robot then learns from both self-execution and active observation, performs quasi-real-time training and adjustment, and consolidates the validated result into the local method library for future reuse. Through this recurring closed-loop process, the robot gradually converts both execution-derived and observation-derived experience into reusable local capability while reducing future dependence on repeated external LLM interaction.
Results show that the proposed framework reduces execution time and LLM dependence in both repeated-task self-execution and observation-driven settings, for example reducing the average total execution time from 7.7772\,s to 6.7779\,s and the average number of LLM calls per task from 1.0 to 0.2 in the repeated-task self-execution experiments.
\end{abstract}

\begin{IEEEkeywords}
    Autonomous Robotics, Active Learning, Large Language Models (LLM), Self-Learning
\end{IEEEkeywords}

\IEEEpeerreviewmaketitle

\section{Introduction}
\label{sec:introduction}

Autonomous robots \cite{misaros2023autonomous} operating in open environments are expected to continuously handle newly encountered tasks rather than only execute a fixed set of predefined behaviors.
However, in many practical systems, a robot can effectively solve a task only when an appropriate method, model, or control procedure has already been prepared in advance.
Once the robot encounters a task that is not covered by its existing local method library, its capability often becomes limited by predefined rules, fixed pipelines, or repeated dependence on external intelligence.

Recent advances in large language models (LLMs) have provided new possibilities for robotic task planning, reasoning, and decision support \cite{wang2025large} \cite{kim2024survey}.
By leveraging broad prior knowledge and flexible instruction understanding, LLMs can help robots decompose tasks, suggest action sequences, and provide high-level guidance in unfamiliar situations.
Nevertheless, directly relying on an LLM whenever an uncovered task appears still has important limitations.
First, repeated interaction with the LLM introduces additional time cost and external dependence during task execution.
Second, even if the robot successfully completes a task with LLM assistance once, the obtained experience is not always transformed into a reusable local method for future tasks.

Meanwhile, even when the robot observes that a human or another external agent has successfully completed a task, such external success is often treated only as a temporary reference rather than being autonomously converted into the robot's own learnable experience.
As a result, the robot may repeatedly solve similar tasks through repeated LLM interaction or repeated external guidance instead of gradually becoming more autonomous.

This limitation becomes more evident in open environments, where tasks are not fully enumerable before deployment.
In such settings, a robot should not only execute tasks, but should also be able to determine whether an existing local method is sufficient, decide when learning should be triggered, organize how a new solution should be learned, and consolidate acquired experience into reusable knowledge.
Such experience may come not only from the robot's own execution, but also from observing successful behaviors of humans or other external agents.
In other words, the robot should not merely solve the current problem; it should also autonomously organize the learning process for previously uncovered tasks based on both execution and observation.

The proactivity considered in this paper is reflected in two aspects.
First, the robot does not wait for a human designer to manually specify the learning procedure for each uncovered task; instead, it autonomously determines whether learning should be triggered, how learning should be organized, and what information should be collected.
Second, the target setting is not a one-shot task-solving scenario in which learning ends after a single execution.
Instead, tasks may recur during long-term robot operation, and learning proceeds alongside task execution so that acquired solutions can be continuously refined and reused in later rounds.
In this paper, such high-level reasoning and judgment are performed by the LLM, although other general-purpose reasoning models could also be used in principle.

To this end, we propose an autonomous learning framework for robots facing uncovered tasks in open environments.
The framework first retrieves the local method library to determine whether a reusable solution already exists for the current task.
If a suitable method is found, the robot directly reuses it.
Otherwise, an LLM-driven learning process is triggered.
In this process, the LLM serves as a high-level planner that helps organize task analysis, candidate model selection, data collection requirements, and execution strategy.
The robot then performs self-execution and, when available, active observation of successful external behaviors to collect task-related data.
Importantly, active observation is not treated as a passive record only; it can also serve as an independent trigger for autonomous learning when the robot detects that an observed task is relevant but not yet covered by its local method library.
After or during task execution, the system performs quasi-real-time training and adjustment, and the validated result is written back into the local method library as a reusable method.
Through this design, the robot gradually converts previously uncovered tasks into locally reusable knowledge.

The main idea of the proposed framework is that the role of the LLM should move from repeated direct dependence at execution time to selective support for learning organization when new tasks appear.
Instead of querying the LLM every time a similar task is encountered, the robot should progressively reduce such interaction by learning and storing reusable methods locally.
This mechanism is especially suitable for repeated or structurally similar tasks, where an initial learning cost can be amortized over later executions.

The main contributions of this paper are summarized as follows:
\begin{itemize}
    \item We formulate the problem of autonomous robot learning for tasks not covered by the local method library in open environments.
    \item We propose an LLM-driven autonomous learning framework that integrates method retrieval, learning triggering, planning, quasi-real-time training, and local knowledge consolidation into a closed loop.
    \item We formalize a two-level notion of proactivity for autonomous robot learning in open environments, where the robot not only organizes learning for uncovered tasks without manual human specification of the learning procedure, but also continues refining and reusing acquired methods through recurring task execution.
\end{itemize}

The remainder of this paper is organized as follows.
Section~\ref{sec:related_work} reviews related studies on LLM-based robotics, continual and open-ended robot learning, and learning from observation.
Section~\ref{sec:model} presents the proposed autonomous learning framework, including local method retrieval, learning triggering, LLM-driven planning, execution and observation-based data acquisition, quasi-real-time training, and knowledge consolidation.
Section~\ref{sec:verification} reports the experimental verification in both self-execution and observation-driven settings.
Finally, Section~\ref{sec:conclusion} concludes the paper and discusses future work.

\section{Related Work}
\label{sec:related_work}

\subsection{LLM-Based Robotics and Task Planning}

Large language models (LLMs) and related foundation models have recently become an important tool for robotic task planning, grounding, and high-level decision making.
Compared with conventional robotic systems that rely heavily on manually designed pipelines, fixed symbolic rules, or narrowly trained task-specific models, LLM-based approaches provide a more flexible interface between language instructions, world knowledge, perception, and action.
This flexibility is especially valuable in open or weakly structured environments, where the robot may encounter tasks that are difficult to fully enumerate before deployment.

Representative systems such as PaLM-E~\cite{driess2023palme}, RT-2~\cite{brohan2023rt2}, and VoxPoser~\cite{huang2023voxposer} demonstrate that language-based models can connect instructions, perception, and action generation for embodied tasks.
PaLM-E shows how a multimodal language model can integrate visual and robotic inputs into a unified reasoning process, thereby supporting embodied decision making across diverse tasks~\cite{driess2023palme}.
RT-2 further illustrates how vision-language-action modeling can transfer semantic knowledge from web-scale data into robotic control, which improves generalization to unseen instructions and novel task compositions~\cite{brohan2023rt2}.
VoxPoser highlights another important direction, namely the use of language models for high-level planning over structured spatial representations, allowing robots to reason about manipulation goals in a more compositional manner~\cite{huang2023voxposer}.
More broadly, recent surveys show that foundation models are increasingly used in robot learning, planning, grounding, and decision support~\cite{xiao2025foundation}.
Taken together, these studies indicate that LLMs provide strong capabilities in semantic reasoning, task decomposition, commonsense guidance, and flexible planning for robots.

Another important contribution of this line of work is that it shifts robotic intelligence from rigid task-specific programming toward instruction-driven and context-aware behavior generation.
Instead of requiring every task to be represented by a fixed handcrafted procedure, LLM-based systems can often infer intermediate subtasks, produce action suggestions, interpret human intent, and adapt high-level plans according to context.
This makes them attractive for embodied agents that must respond to novel commands, incomplete information, or changing environments.
In this sense, LLMs are not merely additional control modules, but increasingly serve as general-purpose reasoning components that connect perception, language, memory, and action.

However, most existing LLM-based robotic systems mainly focus on solving the current task with external model assistance.
Their main objective is typically to improve task execution quality, planning flexibility, or cross-task generalization at inference time.
As a result, they often treat the LLM as a repeatedly queried external capability.
Even when an LLM successfully helps the robot complete a previously uncovered task, the obtained result is not always transformed into reusable local knowledge for future tasks.
In many cases, the system remains dependent on the LLM whenever a similar task reappears, which limits long-term autonomy and may increase execution latency, interaction overhead, and external dependence.

This limitation becomes even more important in open environments, where robots are expected not only to solve unfamiliar tasks once, but also to gradually accumulate reusable capability over time.
From this perspective, the key challenge is no longer only how to use an LLM to generate a good plan for the present task, but also how to determine when LLM support should be invoked, how the resulting experience should be organized into a learnable form, and how the learned result should be written back into the robot's own local method repository.
Existing LLM-guided robotic systems only partially address this broader problem.

By contrast, our work uses the LLM as a selective high-level planner that is invoked when the local method library does not cover the current task, rather than as a default dependency for repeated execution.
Moreover, the role of the LLM in our framework is not limited to producing an immediate action sequence.
Instead, it helps organize the learning process itself, including task analysis, candidate model selection, data collection planning, and subsequent method formation.
The validated result is then consolidated into the local method library for future reuse.
Therefore, compared with prior LLM-guided robotic systems, our framework places stronger emphasis on reducing future dependence on repeated LLM interaction through local method consolidation and autonomous capability accumulation.

\subsection{Continual and Open-Ended Robot Learning}

Continual, lifelong, and open-ended robot learning aim to enable robots to accumulate capability over time instead of learning each task in isolation.
Compared with conventional single-task training settings, these directions emphasize that a robot deployed in the real world should continuously adapt to new tasks, environments, and requirements while preserving useful prior knowledge.
This problem is especially important for long-term autonomous systems, because practical environments are rarely static and the full range of future tasks is typically unknown before deployment.

A major motivation behind continual robot learning is that repeated retraining from scratch is inefficient and does not reflect how an autonomous agent should grow after deployment.
Instead, the robot is expected to retain useful knowledge acquired from previous tasks, transfer it to related new tasks, and gradually expand its effective capability boundary.
Recent surveys on continual reinforcement learning highlight key challenges such as catastrophic forgetting, positive and negative transfer, changing task distributions, and sequential adaptation under limited memory and computation budgets~\cite{pan2025crl}.
Related studies on robotic lifelong learning further investigate how knowledge can be preserved, reused, and combined across long interaction horizons, thereby supporting sustained performance improvement over time~\cite{meng2025lifelong}.

Another important contribution of this research direction is that it shifts the evaluation focus from one-time task success to long-term capability growth.
In other words, the quality of a learning system is no longer judged only by whether it can solve the current task, but also by whether it can preserve useful knowledge, reduce future relearning cost, and improve adaptability to subsequent tasks.
This perspective is closely related to the needs of open-environment robotics, where the value of a system lies not only in immediate task completion, but also in its ability to accumulate reusable competence under evolving conditions.

However, many existing continual or lifelong learning settings still assume relatively structured learning scenarios.
For example, the system may be given a pre-defined stream of tasks, a known adaptation mechanism, a fixed policy update rule, or a relatively clear boundary between training and testing phases.
Even when the task sequence is long, the learning pipeline itself is often largely predefined.
As a result, these methods mainly address how to update model parameters or preserve task knowledge over time, rather than how the robot should autonomously decide when learning should be triggered, what kind of learning process should be organized, and how the resulting solution should be represented as reusable local task knowledge.

This distinction is important for the problem studied in this paper.
In our setting, the robot may encounter a task that is not covered by its current local method library, and the first challenge is not merely how to adapt parameters, but how to determine that the task is uncovered, whether direct reuse is possible, and whether a new learning process should be initiated at all.
Moreover, once learning is triggered, the robot must organize not only model updating, but also task analysis, candidate model selection, data collection planning, and method consolidation.
Therefore, compared with standard continual learning, the issue here is not only continuous adaptation, but also the autonomous organization of learning for previously uncovered tasks in open environments.

In addition, many continual learning methods primarily focus on improving a model or policy representation, whereas our work places stronger emphasis on the accumulation of reusable \emph{methods}.
That is, the output of learning in our framework is not only an updated parameter vector, but also a locally reusable problem-solving unit that includes task-relevant procedures, learned configurations, and applicability information.
This design is particularly useful for reducing future dependence on repeated external support, because once a task has been learned and consolidated, the robot can directly retrieve and reuse the corresponding local method instead of repeatedly invoking external intelligence.

By contrast, our framework addresses a complementary but more operational question in open-environment robotics:
how a robot should move from encountering an uncovered task to triggering learning, organizing learning, and finally converting the acquired result into reusable local capability.
Therefore, compared with prior continual and open-ended robot learning research, our work places stronger emphasis on local method retrieval, learning triggering, and knowledge consolidation as part of a unified autonomous learning loop.

\subsection{Learning from Observation and Demonstration}

Another closely related direction is robot learning from observation, demonstration, and human video.
Recent surveys show that learning from video or demonstration can exploit rich external behavior without requiring direct robot action labels, making it attractive for scalable robot learning~\cite{eze2024videoreview}.
Related works further explore learning from large-scale human videos or external successful behaviors to improve robot generalization and transfer~\cite{mao2024humanoidx}.
These efforts suggest that external successful behavior can serve as a valuable source of learnable structure for robots.

However, most existing methods in this direction mainly focus on imitation or policy transfer from demonstrations to execution.
They usually do not address the higher-level question of when a robot should autonomously decide to learn from an observed successful event, nor how such observed experience should be integrated with LLM-based planning and stored as a reusable local method within a broader autonomous learning loop.
In contrast, our framework explicitly treats observation as one of the sources for learning triggering and planning, alongside self-execution, and emphasizes the consolidation of learned results into a reusable local method library.

Although substantial progress has been made in LLM-guided robotics, continual robot learning, and learning from observation, existing studies do not fully address the combined problem considered in this paper:
how a robot in an open environment should detect that a task is not covered by its local method library, selectively invoke high-level planning support, learn from both execution and observation, and finally convert the acquired result into reusable local knowledge that reduces future dependence on repeated external LLM interaction.
This gap motivates the framework proposed in this work.

\section{Uncovered-Task-Centered Local Reuse Learning Model}
\label{sec:model}

This section presents the proposed autonomous learning framework for robots facing tasks that are not covered by the local method library in open environments.
The core objective of the model is to answer three key questions in a unified manner:
\emph{when learning should be triggered, how learning should be planned, and how newly acquired experience should be consolidated into reusable knowledge}.
To this end, the proposed model organizes robot behavior into a closed loop including method retrieval, learning triggering, LLM-driven planning, execution and data collection, quasi-real-time training, and method consolidation.

\subsection{Overview of the Model}
\label{subsec:overview}

The proposed model assumes that a robot operates in an open environment and continuously encounters tasks, events, or problem situations.
For each encountered task, the robot first checks whether the local method library already contains an applicable solution.
If a suitable method is found, the robot directly reuses it.
Otherwise, the robot activates an autonomous learning process to generate, improve, and store a new solution.

Different from conventional task-specific learning pipelines, the proposed model does not assume that the task category, model type, data requirement, or training procedure is fully predefined.
Instead, these elements are planned dynamically according to the current task characteristics and the robot's existing knowledge.
A large language model (LLM) is used as the high-level planning component to organize the learning process, while the robot itself performs task execution, data collection, model training, and knowledge accumulation.

Accordingly, the proposed model contains the following major components:

\begin{itemize}
    \item \textbf{Task perception and problem formulation module}, which detects a task or problem that requires handling;
    \item \textbf{Local method retrieval module}, which checks whether an existing reusable method is available;
    \item \textbf{Learning trigger module}, which decides whether autonomous learning should be activated;
    \item \textbf{LLM-driven learning planning module}, which determines what should be learned, what model should be used, and what data should be collected;
    \item \textbf{Execution and data acquisition module}, which performs actions or observations and records the resulting data;
    \item \textbf{Quasi-real-time training and adjustment module}, which updates the task-solving model immediately after or during execution;
    \item \textbf{Knowledge consolidation module}, which stores validated new methods into the local method library.
\end{itemize}

These modules jointly form the core processing chain for a currently encountered task.
To summarize how the proposed framework handles one task episode from perception to local method update, Algorithm~\ref{alg:overview_model} presents the single-task workflow of the proposed autonomous learning model.
The broader long-term recurring operation of the framework is further described later in Section~\ref{subsec:closed_loop}.

\begin{algorithm}[t]
\caption{Single-Task Workflow of the Proposed Autonomous Learning Model}
\label{alg:overview_model}
\KwIn{Current task $T$, local method library $\mathcal{M}$, historical information $\mathcal{H}$}
\KwOut{Executed solution and updated local method library $\mathcal{M}$}

Perceive task and construct task descriptor for $T$\;

Retrieve the best matching method $M^*$ from $\mathcal{M}$\;

\eIf{$M^*$ exists and is reliable for $T$}{
    Execute $M^*$ directly\;
    Record execution result and update its performance record\;
}{
    Trigger autonomous learning\;

    Use the LLM to generate learning plan
    $\Pi = \Phi_{\text{LLM}}(T,\mathcal{H},\mathcal{F})$\;

    Determine candidate model structure, required data, and execution/observation strategy from $\Pi$\;

    Execute actions and/or observe external behaviors to collect task data $\mathcal{D}_{\text{task}}$\;

    Perform quasi-real-time adjustment during execution if intermediate feedback is available\;

    Train or refine the task model using $\mathcal{D}_{\text{task}}$\;

    Validate the obtained solution and construct a new reusable method $M_{\text{new}}$\;

    Store $M_{\text{new}}$ into the local method library:
    $\mathcal{M} \leftarrow \mathcal{M} \cup \{M_{\text{new}}\}$\;

    Execute the validated new method if needed\;
}
Return execution result and updated $\mathcal{M}$\;
\end{algorithm}

Algorithm~\ref{alg:overview_model} shows how the proposed framework handles a single currently encountered task.
The workflow starts from task perception and method retrieval.
If a suitable method already exists in the local library, the robot reuses it and updates its historical performance record.
This branch ensures that previously learned knowledge can be efficiently reused instead of repeatedly relearning similar tasks.

If no reliable method is found, the framework enters the autonomous learning branch.
In this branch, the LLM acts as a high-level planner that organizes the learning process rather than directly replacing the task execution module.
Specifically, it determines the key subproblems, suggests candidate model structures, specifies the data that should be collected, and arranges the execution or observation strategy.
Based on this plan, the robot interacts with the environment to collect task-relevant data through self-execution and, when available, active observation of external agents or events.

After data collection begins, the robot can perform quasi-real-time adjustment according to intermediate feedback, which allows the current solution to be refined before the entire learning cycle finishes.
Once sufficient data have been obtained, the task-solving model is trained or updated, and the resulting solution is validated.
If the solution is considered effective, it is transformed into a reusable method and stored in the local method library.
In this way, the framework realizes a complete task-level learning cycle from task encounter to local knowledge update.

This single-task workflow also clarifies the role of the LLM in the proposed framework.
Traditional LLM-guided robotic systems often rely on repeated external assistance at execution time.
By contrast, in our framework, the LLM is primarily used to organize learning for currently uncovered tasks, while the validated result is written back into the robot's own local method library.
The long-term implication of this design is further captured by the recurring closed-loop process described later, in which the output of one task-level learning cycle becomes reusable knowledge for future tasks.

\subsection{Task Representation and Method Library}
\label{subsec:task_library}

Let the current task be denoted by $T$.
Each task is represented by a task descriptor
\begin{equation}
T = \{G, E, O, C\},
\end{equation}
where $G$ denotes the task goal, $E$ denotes the environmental context, $O$ denotes the available observations, and $C$ denotes the execution constraints.

The robot maintains a local method library
\begin{equation}
\mathcal{M} = \{M_1, M_2, \dots, M_N\},
\end{equation}
where each stored method $M_i$ contains not only an executable solution but also its applicability conditions.
A method can be represented as
\begin{equation}
M_i = \{P_i, \Theta_i, D_i, A_i, R_i\},
\end{equation}
where $P_i$ is the procedural description, $\Theta_i$ denotes model parameters or policy parameters, $D_i$ is the associated data profile, $A_i$ represents applicability conditions, and $R_i$ records historical performance or reliability.

Given a task $T$, the retrieval module searches for a matching method
\begin{equation}
M^* = \arg\max_{M_i \in \mathcal{M}} S(T, M_i),
\end{equation}
where $S(\cdot,\cdot)$ denotes the matching score between the current task and a stored method.
If the best score satisfies
\begin{equation}
S(T, M^*) \geq \tau_r,
\end{equation}
where $\tau_r$ is a retrieval threshold, the robot directly reuses $M^*$.
Otherwise, the task is regarded as \emph{not covered} by the current local method library, and autonomous learning is triggered.

\subsection{Learning Trigger Mechanism}
\label{subsec:trigger}

The learning trigger mechanism determines whether the robot should start a new learning process.
The trigger decision is based on method coverage, execution confidence, and observed failure risk.
In addition to task-driven triggering, the proposed framework also supports observation-driven triggering.
That is, when the robot observes that a human or another external agent successfully completes a task or subtask that is relevant to its future capability, but the corresponding method is not yet covered by the local method library, the robot may proactively trigger learning even without having executed the task by itself.
In this way, successful external observation is treated not merely as temporary guidance, but as a potential source for autonomous capability acquisition.

Let the trigger indicator be denoted by $z \in \{0,1\}$, where $z=1$ means that learning is triggered.
The trigger decision considers not only uncovered-task execution and insufficient method confidence, but also successful external observation that reveals a potentially useful yet uncovered capability.
The decision rule is formulated as
\begin{equation}
z =
\begin{cases}
1, & S(T, M^*) < \tau_r, \\
1, & S(T, M^*) \geq \tau_r \ \text{and}\ Q(M^*,T) < \tau_q, \\
1, & O_{\mathrm{succ}} = 1 \ \text{and}\ S(T_{\mathrm{obs}}, M^*_{\mathrm{obs}}) < \tau_o, \\
0, & \text{otherwise},
\end{cases}
\end{equation}
where $Q(M^*,T)$ denotes the expected suitability or confidence of applying the retrieved method to task $T$, $\tau_q$ is the confidence threshold, $O_{\mathrm{succ}}$ indicates whether a successful external observation has been detected, $T_{\mathrm{obs}}$ denotes the observed task descriptor, $M^*_{\mathrm{obs}}$ is the best matching local method for the observed task, and $\tau_o$ is the observation-trigger threshold.

This design allows the robot to trigger learning not only when no matching method exists, but also when an existing method is deemed insufficient for reliable execution, or when successful external observation reveals a relevant capability that has not yet been internalized.
Therefore, the framework supports \emph{uncovered-task learning}, \emph{insufficient-method refinement}, and \emph{observation-driven proactive learning}.

\subsection{LLM-Driven Learning Planning}
\label{subsec:planning}

Once learning is triggered, the LLM-driven planning module organizes the learning process.
Its role is not to directly replace low-level execution, but to determine the learning strategy for the current task.

Given the task descriptor $T$, the retrieved historical information $\mathcal{H}$, and current feedback $\mathcal{F}$, the LLM produces a planning output
\begin{equation}
\Pi = \Phi_{\text{LLM}}(T, \mathcal{H}, \mathcal{F}),
\end{equation}
where $\Phi_{\text{LLM}}(\cdot)$ denotes the planning function driven by the LLM.
The planning result $\Pi$ includes at least the following elements:
\begin{equation}
\Pi = \{K, \mathcal{N}, \mathcal{D}, \mathcal{A}, \mathcal{U}\},
\end{equation}
where $K$ denotes the key subproblems to solve, $\mathcal{N}$ denotes the candidate model structure or model combination, $\mathcal{D}$ denotes the required data to collect, $\mathcal{A}$ denotes the execution or observation strategy, and $\mathcal{U}$ denotes the update criteria.

The candidate model set $\mathcal{N}$ may include sequential models, visual models, multimodal models, or their combinations, such as LSTM, Transformer, CNN, or hybrid architectures.
For instance, if the task mainly involves action sequences, the LLM may assign a higher priority to sequence modeling.
If the task relies more on visual cues, visual perception models may be selected.
Thus, the planning process connects task characteristics with model selection and data requirements.

\subsection{Execution and Data Acquisition}
\label{subsec:execution_data}

After a learning plan is generated, the robot interacts with the environment to collect the data required for model construction or refinement.
The proposed framework supports two complementary experience sources: self-execution and active observation.
Self-execution corresponds to the robot's own task attempts, while active observation corresponds to observing successful behaviors of humans or other external agents.
Unlike conventional settings in which observation is treated only as auxiliary context, the proposed framework regards active observation as a valid source of learnable task knowledge that can independently support model construction and method acquisition.

Let the collected data at time step $t$ be represented as
\begin{equation}
x_t = \{o_t, a_t, r_t, e_t\},
\end{equation}
where $o_t$ is the observation, $a_t$ is the executed or observed action, $r_t$ is the resulting outcome or reward-like feedback, and $e_t$ denotes contextual information such as environmental state or external guidance.

Over one task episode, the collected dataset is
\begin{equation}
\mathcal{D}_{\text{task}} = \{x_1, x_2, \dots, x_T\}.
\end{equation}

The overall acquired data can be written as
\begin{equation}
\mathcal{D} = \mathcal{D}_{\text{self}} \cup \mathcal{D}_{\text{obs}},
\end{equation}
where $\mathcal{D}_{\text{self}}$ is collected from the robot's own execution and $\mathcal{D}_{\text{obs}}$ is collected from observing external agents, humans, or environmental processes.

Here, $\mathcal{D}_{\text{obs}}$ is not limited to passive records of surrounding events.
Instead, it represents structured observation data extracted from successful external behaviors, including observed action sequences, contextual conditions, and resulting outcomes.
Such data can be used not only to refine an existing method, but also to initialize or construct a new local method for a previously uncovered task.
Therefore, observation is treated as an independent learning source rather than merely a supplementary signal.

This mechanism enables the robot to learn not only from its own task attempts but also from successfully observed external behaviors.
As a result, the framework broadens the source of learnable experience in open environments and allows the robot to proactively transform both execution-derived and observation-derived information into reusable local capability.

\subsection{Quasi-Real-Time Training and Adjustment}
\label{subsec:training}

A key property of the proposed framework is quasi-real-time learning.
This means that data collection, feedback analysis, and model update are closely coupled with task execution rather than being postponed to a separate long offline phase.

Let the task model selected by the planner be denoted by $f_{\theta}$, where $\theta$ represents trainable parameters.
Given the task data $\mathcal{D}_{\text{task}}$, the model is updated by minimizing
\begin{equation}
\theta^{*} = \arg\min_{\theta} \mathcal{L}(f_{\theta}; \mathcal{D}_{\text{task}}),
\end{equation}
where $\mathcal{L}(\cdot)$ is a task-dependent training loss.

The quasi-real-time property can be expressed by dividing training into two stages:
\begin{equation}
\theta \rightarrow \theta' \rightarrow \theta^{*},
\end{equation}
where $\theta'$ denotes intermediate adjustment during or immediately after execution, and $\theta^{*}$ denotes the refined model after the full episode update.

This design avoids a purely delayed learning mode and allows the robot to rapidly improve the solution quality for similar future tasks.
Moreover, intermediate results can be sent back to the LLM to revise the current plan:
\begin{equation}
\Pi' = \Phi_{\text{LLM}}(T, \mathcal{H}, \mathcal{F}'),
\end{equation}
where $\mathcal{F}'$ denotes updated intermediate feedback.
Therefore, planning and training form a bidirectional interaction.

\subsection{Knowledge Consolidation and Method Update}
\label{subsec:consolidation}

After training and validation, the newly obtained task-solving strategy is transformed into a reusable method and stored in the local method library.
Let the learned new method be denoted by $M_{\text{new}}$.
Then the library update process is
\begin{equation}
\mathcal{M} \leftarrow \mathcal{M} \cup \{M_{\text{new}}\}.
\end{equation}

The new method is not limited to raw model parameters.
Instead, it includes executable steps, model configuration, data characteristics, and applicability conditions:
\begin{equation}
M_{\text{new}} = \{P_{\text{new}}, \Theta_{\text{new}}, D_{\text{new}}, A_{\text{new}}, R_{\text{new}}\}.
\end{equation}

To avoid repeatedly using a partially defective but executable solution, the robot can also maintain a refinement mechanism.
Let $U(M)$ denote the long-term utility of a method.
Then method refinement can be triggered when
\begin{equation}
U(M_{\text{new}}) < \tau_u,
\end{equation}
where $\tau_u$ is the utility threshold.
In this case, the system may re-enter the planning and improvement process to search for a simpler, more robust, or more efficient solution.

\subsection{Closed-Loop Learning Process}
\label{subsec:closed_loop}

The above modules describe how the proposed framework handles a task when it is encountered.
In particular, Algorithm~\ref{alg:overview_model} summarizes the overall workflow for a current task, including method retrieval, learning triggering, planning, execution or observation, training, and knowledge consolidation.
However, the target setting of this paper is not a one-shot task-solving scenario in which learning ends after a single task attempt.
Instead, the robot is assumed to operate continuously in an open environment, where tasks may recur, related subtasks may reappear, and newly acquired capability should influence later executions.
Therefore, beyond the single-task workflow, the overall behavior of the proposed system should be understood as a long-term closed-loop autonomous learning process.

At a high level, this closed loop repeatedly performs the following operations: perceiving the current task, checking whether existing local methods are sufficient, triggering learning when necessary, organizing the learning process through high-level reasoning, collecting execution- or observation-derived experience, updating local capability, and reusing the consolidated result in later tasks.
To make this long-term mechanism explicit, Algorithm~\ref{alg:closed_loop_process} summarizes the recurring closed-loop learning process of the proposed framework.

\begin{algorithm}[t]
\caption{Closed-Loop Autonomous Learning over Recurring Tasks}
\label{alg:closed_loop_process}
\KwIn{Initial local method library $\mathcal{M}$, reasoning model $\Phi$ (LLM in this paper)}
\KwOut{Continuously updated local method library $\mathcal{M}$}

\While{the robot remains in operation}{
    perceive a current task, event, or observed successful external behavior\;

    formulate the corresponding task descriptor $T$\;

    retrieve the best matching local method $M^*$ from $\mathcal{M}$\;

    \uIf{a reliable method exists for $T$}{
        execute or reuse $M^*$ directly\;
        record execution outcome and update method statistics\;
    }
    \Else{
        trigger autonomous learning\;

        use the reasoning model $\Phi$ to organize learning, including task analysis, candidate model selection, data requirement identification, and execution/observation strategy planning\;

        perform self-execution and/or active observation to collect task-related data\;

        conduct quasi-real-time adjustment and post-episode training\;

        validate the obtained solution and construct a reusable local method $M_{\text{new}}$\;

        update the local method library:
        $\mathcal{M} \leftarrow \mathcal{M} \cup \{M_{\text{new}}\}$\;
    }

    continue to the next encountered task or observed event\;
}
\end{algorithm}

Algorithm~\ref{alg:closed_loop_process} emphasizes that the proposed framework is not limited to a single isolated task episode.
Instead, it continually alternates between local reuse and new learning as the robot remains in operation.
Once a previously uncovered task has been learned, the resulting method becomes part of the robot's local capability and can be reused in later executions.
Therefore, the output of one learning cycle is not the end of the process, but the input knowledge of later cycles.

This recurring process can be summarized at a high level as
\begin{equation}
(T_t, \mathcal{M}_t) \rightarrow (M_t^*, z_t, M_t^{\mathrm{new}}) \rightarrow \mathcal{M}_{t+1},
\end{equation}
where $T_t$ denotes the currently encountered task or observed event at cycle $t$, $\mathcal{M}_t$ denotes the local method library at that time, $M_t^*$ denotes the retrieved best-matching method, $z_t \in \{0,1\}$ indicates whether autonomous learning is triggered, and $M_t^{\mathrm{new}}$ denotes the validated new method obtained in the current cycle.

Accordingly, the local method library evolves over time as
\begin{equation}
\mathcal{M}_{t+1} =
\begin{cases}
\mathcal{M}_t, & \text{if no new method},\\
\mathcal{M}_t \cup \{M_t^{\mathrm{new}}\}, & \text{otherwise}.
\end{cases}
\end{equation}
This formulation emphasizes that the output of one learning cycle becomes part of the robot's reusable local capability in later cycles.

The proactivity of the proposed framework is reflected in two aspects.
First, the robot does not wait for a human designer to manually specify the learning procedure for each uncovered task.
Instead, it autonomously determines whether learning should be triggered, how learning should be organized, and what information should be collected.
In this paper, such high-level reasoning and judgment are performed by the LLM, although other general-purpose reasoning models could also be used in principle.

Second, the target setting considered here is not a one-time execution setting in which the learning process terminates after one attempt.
Rather, the framework is designed for recurring task settings in open environments, where similar or related tasks may reappear during long-term robot operation.
Accordingly, learning is carried out in parallel with task execution and continues to refine local capability over repeated task cycles.
In this sense, the proposed framework is not merely an execution loop, but a proactive learning loop that continuously organizes, updates, and reuses the robot's own task-solving methods.

This closed-loop design enables the robot to continuously expand its solvable task scope in open environments.
Instead of relying only on predefined methods or repeatedly depending on external assistance, the robot gradually accumulates reusable knowledge through repeated interaction, learning, observation, and consolidation.







\section{Theoretical Analysis}
\label{sec:theory}

This section provides a theoretical analysis of the proposed framework from two levels.
First, we analyze the rationality of the core mechanism, including method retrieval, learning triggering, LLM-driven planning, and knowledge consolidation.
Second, we analyze the cost structure and long-term benefit of the framework.
The purpose of this section is not to prove global optimality under overly restrictive assumptions, but to show why the proposed design is theoretically reasonable for robots facing tasks that are not covered by the local method library in open environments.

\subsection{Mechanism-Level Rationality Analysis}
\label{subsec:mechanism_analysis}

The proposed framework is built on the principle that a robot should not always relearn every encountered task from scratch, nor should it completely rely on a fixed predefined method set.
Instead, it should first attempt to reuse existing knowledge and trigger learning only when existing knowledge is missing or unreliable.
Below, we analyze the main properties of this mechanism.

\subsubsection{Property 1: Retrieval before learning avoids unnecessary relearning}
\label{subsubsec:property1}

Let $T$ denote the current task, and let $\mathcal{M}$ denote the local method library.
Suppose that there exists a stored method $M^* \in \mathcal{M}$ such that $M^*$ is applicable to $T$ and can solve $T$ with satisfactory quality.
If the retrieval module correctly identifies $M^*$, then the robot can directly execute it without invoking a new learning process.

Let the cost of reusing an existing method be denoted by
\begin{equation}
C_{\text{reuse}} = C_{\text{retrieve}} + C_{\text{exec}},
\end{equation}
where $C_{\text{retrieve}}$ is the method retrieval cost and $C_{\text{exec}}$ is the execution cost.

If the robot instead relearns the task from scratch, the total cost becomes
\begin{equation}
C_{\text{learn}} = C_{\text{retrieve}} + C_{\text{plan}} + C_{\text{collect}} + C_{\text{train}} + C_{\text{exec}},
\end{equation}
where $C_{\text{plan}}$, $C_{\text{collect}}$, and $C_{\text{train}}$ denote the planning, data collection, and training costs, respectively.

Since learning-related costs are nonnegative, we have
\begin{equation}
C_{\text{learn}} - C_{\text{reuse}} = C_{\text{plan}} + C_{\text{collect}} + C_{\text{train}} \geq 0.
\end{equation}

Therefore, whenever a reusable method already exists and can be correctly identified, reusing it is never more expensive than relearning the same task from scratch.
This property justifies the design choice that method retrieval should precede autonomous learning.

\subsubsection{Property 2: Learning triggering is necessary for uncovered or unreliable tasks}
\label{subsubsec:property2}

A robot operating in an open environment may encounter tasks for which the local method library contains no suitable solution, or only contains a weakly related but unreliable method.
If the framework never triggers learning in such situations, then the robot can only rely on inappropriate reuse or direct failure.

Let $S(T,M^*)$ denote the matching score between the current task $T$ and the best retrieved method $M^*$, and let $Q(M^*,T)$ denote the expected suitability of applying $M^*$ to $T$.
The proposed framework triggers learning when either
\begin{equation}
S(T,M^*) < \tau_r
\end{equation}
or
\begin{equation}
Q(M^*,T) < \tau_q,
\end{equation}
where $\tau_r$ and $\tau_q$ are predefined thresholds.

This mechanism has two rational consequences.

First, if no method satisfies the coverage requirement, then autonomous learning is necessary because direct reuse does not exist as a valid solution path.

Second, even if a roughly matching method exists, but its expected suitability is low, then triggering learning can prevent the robot from repeatedly applying a partially defective solution.
Hence, the trigger mechanism is not only a missing-method detector, but also a safeguard against low-quality reuse.

\subsubsection{Property 3: LLM-driven planning reduces blind trial-and-error}
\label{subsubsec:property3}

Consider the case where a task is not covered by the local method library and a learning process must be started.
Without a planning mechanism, the robot may choose model types, data collection strategies, and execution procedures arbitrarily.
Such blind trial-and-error typically increases both the amount of collected data and the number of failed attempts.

Let $\Omega$ denote the space of candidate learning configurations, where each configuration includes model structure, data collection plan, and execution strategy.
Suppose that only a subset $\Omega^{+} \subseteq \Omega$ is reasonably suitable for the current task.
If the robot selects a configuration uniformly at random, the probability of selecting a suitable configuration is
\begin{equation}
P_{\text{rand}} = \frac{|\Omega^{+}|}{|\Omega|}.
\end{equation}

Now suppose that the LLM-driven planner provides a filtered candidate set $\Omega_{\text{LLM}} \subseteq \Omega$ such that
\begin{equation}
\Omega^{+} \cap \Omega_{\text{LLM}}
\end{equation}
contains a larger proportion of suitable configurations than the original search space.
Then the probability of selecting a suitable configuration becomes
\begin{equation}
P_{\text{LLM}} = \frac{|\Omega^{+} \cap \Omega_{\text{LLM}}|}{|\Omega_{\text{LLM}}|}.
\end{equation}

If the planner is informative, then typically
\begin{equation}
P_{\text{LLM}} \geq P_{\text{rand}}.
\end{equation}

This means that the LLM-driven planning mechanism can improve the chance that the robot starts learning from a more appropriate configuration, thereby reducing blind exploration.
Although this does not guarantee optimal planning, it theoretically explains why a task-aware planner is preferable to uninformed selection.

\subsubsection{Property 4: Knowledge consolidation reduces future solving cost for similar tasks}
\label{subsubsec:property4}

After a new task has been solved through autonomous learning, the resulting solution is converted into a reusable method and stored in the local method library.
This creates the possibility of direct reuse in future similar tasks.

Let $T_1$ be an uncovered task encountered at time step $1$.
After learning, the system generates a new method $M_{\text{new}}$ and updates the method library:
\begin{equation}
\mathcal{M} \leftarrow \mathcal{M} \cup \{M_{\text{new}}\}.
\end{equation}

Now consider a future task $T_2$ such that $T_2$ is similar enough to $T_1$ that $M_{\text{new}}$ becomes retrievable and applicable.
Without consolidation, the robot would need to repeat the full learning cost:
\begin{equation}
C_{\text{future,no-store}} = C_{\text{retrieve}} + C_{\text{plan}} + C_{\text{collect}} + C_{\text{train}} + C_{\text{exec}}.
\end{equation}
With consolidation, the future cost becomes
\begin{equation}
C_{\text{future,store}} = C_{\text{retrieve}} + C_{\text{exec}}.
\end{equation}

Thus,
\begin{equation}
C_{\text{future,no-store}} - C_{\text{future,store}}
= C_{\text{plan}} + C_{\text{collect}} + C_{\text{train}} \geq 0.
\end{equation}

Therefore, once a newly learned solution becomes reusable, storing it can reduce the expected solving cost for future related tasks.
This is the theoretical basis for the proposed knowledge consolidation mechanism.

\subsubsection{Property 5: Active observation enlarges the experience source}
\label{subsubsec:property5}

If a robot can learn only from self-execution, then its available training data are limited to the states and actions that it has personally experienced.
In open environments, however, useful information may also come from observing humans, other robots, or environmental processes.

Let the total collected data be
\begin{equation}
\mathcal{D} = \mathcal{D}_{\text{self}} \cup \mathcal{D}_{\text{obs}},
\end{equation}
where $\mathcal{D}_{\text{self}}$ is data collected from self-execution and $\mathcal{D}_{\text{obs}}$ is data collected from active observation.

Clearly,
\begin{equation}
|\mathcal{D}| \geq |\mathcal{D}_{\text{self}}|.
\end{equation}

If some useful state-action-outcome patterns appear only in $\mathcal{D}_{\text{obs}}$, then relying only on self-execution would fail to exploit these patterns.
Thus, active observation can broaden the experience distribution available to the learning module, which in turn may reduce the amount of risky or costly self-trial needed.
This property justifies the inclusion of active observation as a complementary learning source.

\subsection{Cost and Long-Term Benefit Analysis}
\label{subsec:cost_analysis}

We next analyze the cost structure of the proposed framework and its long-term benefit.
The key idea is that the framework may incur additional cost when learning is triggered, but this cost is an investment that can reduce future task-solving cost through knowledge reuse.

\subsubsection{Single-task cost decomposition}
\label{subsubsec:single_task_cost}

Let $z \in \{0,1\}$ be the learning trigger indicator for task $T$, where $z=1$ means that autonomous learning is triggered and $z=0$ means that an existing method is directly reused.
Then the total cost of handling a single task can be written as
\begin{equation}
C(T) =
C_{\text{retrieve}} + C_{\text{exec}} +
z \left(
C_{\text{plan}} + C_{\text{collect}} + C_{\text{train}} + C_{\text{store}}
\right),
\label{eq:single_task_cost}
\end{equation}
where $C_{\text{store}}$ denotes the cost of validating and storing a new method.

Equation~(\ref{eq:single_task_cost}) explicitly shows that the framework separates two modes:

\begin{itemize}
    \item \textbf{Reuse mode} ($z=0$): the robot only retrieves and executes;
    \item \textbf{Learning mode} ($z=1$): the robot additionally pays the costs of planning, data collection, training, and storage.
\end{itemize}

This decomposition clarifies that the proposed framework does not always pay the full learning cost.
Instead, learning cost is paid only when the current task is not adequately covered by existing knowledge.

\subsubsection{Expected cost under task streams}
\label{subsubsec:expected_cost}

Consider a sequence of tasks
\begin{equation}
\{T_1, T_2, \dots, T_N\}.
\end{equation}
The total handling cost is
\begin{equation}
C_{\text{total}}(N) = \sum_{i=1}^{N} C(T_i),
\end{equation}
and the average cost per task is
\begin{equation}
\bar{C}(N) = \frac{1}{N} \sum_{i=1}^{N} C(T_i).
\end{equation}

Let $p_i$ denote the probability that task $T_i$ is already covered by the local method library at the time when it is encountered.
Then the expected single-task cost can be written as
\begin{equation}
\mathbb{E}[C(T_i)]
=
C_{\text{retrieve}} + C_{\text{exec}}
+ (1-p_i)\left(C_{\text{plan}} + C_{\text{collect}} + C_{\text{train}} + C_{\text{store}}\right).
\label{eq:expected_single_cost}
\end{equation}

Equation~(\ref{eq:expected_single_cost}) indicates that the expected extra learning cost is weighted by the uncovered-task probability $(1-p_i)$.
As the method library grows and becomes more informative, one expects $p_i$ to increase for future tasks that are similar to previously solved tasks.
Consequently, the expected task-solving cost decreases.

\subsubsection{Long-term reuse benefit}
\label{subsubsec:reuse_benefit}

Suppose that solving one uncovered task generates a reusable method whose future reuse probability is $\rho$, where $0 \leq \rho \leq 1$.
Each successful reuse avoids the repeated cost
\begin{equation}
\Delta C = C_{\text{plan}} + C_{\text{collect}} + C_{\text{train}}.
\end{equation}

If the expected number of future reusable occasions for this method is $K$, then the expected cumulative reuse benefit is
\begin{equation}
B_{\text{reuse}} = \rho K \Delta C.
\end{equation}

Let the one-time investment for producing and storing the method be
\begin{equation}
I = C_{\text{plan}} + C_{\text{collect}} + C_{\text{train}} + C_{\text{store}}.
\end{equation}

Then the net expected long-term benefit is
\begin{equation}
B_{\text{net}} = B_{\text{reuse}} - I
= \rho K \Delta C - \left(C_{\text{plan}} + C_{\text{collect}} + C_{\text{train}} + C_{\text{store}}\right).
\end{equation}

Thus, knowledge consolidation is beneficial in expectation when
\begin{equation}
\rho K \left(C_{\text{plan}} + C_{\text{collect}} + C_{\text{train}}\right)
>
C_{\text{plan}} + C_{\text{collect}} + C_{\text{train}} + C_{\text{store}}.
\label{eq:benefit_condition}
\end{equation}

Equation~(\ref{eq:benefit_condition}) provides a clear interpretation:
if a newly learned method is likely to be reused often enough in the future, then the one-time learning and storage cost is justified.
This is exactly the long-term motivation behind the proposed ``learn and consolidate'' strategy.

\subsubsection{Monotonic tendency of average cost under increasing coverage}
\label{subsubsec:monotonicity}

Assume that the average method coverage probability does not decrease over time, that is,
\begin{equation}
p_{i+1} \geq p_i
\quad \text{for } i = 1,2,\dots,N-1.
\end{equation}
This assumption is reasonable when the task stream contains recurring or related tasks and the method library is continuously updated with validated new solutions.

From Equation~(\ref{eq:expected_single_cost}), if the non-retrieval cost terms remain bounded, then increasing $p_i$ implies
\begin{equation}
\mathbb{E}[C(T_{i+1})] \leq \mathbb{E}[C(T_i)].
\end{equation}

Therefore, under repeated exposure to related tasks, the expected task-solving cost has a non-increasing tendency.
This does not claim strict monotonic decrease for every individual task, since some later tasks may still be entirely novel.
However, it supports the claim that the framework has the potential to reduce long-term average cost as reusable knowledge accumulates.

\subsubsection{Interpretation of quasi-real-time training}
\label{subsubsec:qrr_interpretation}

The proposed framework adopts quasi-real-time training rather than fully delayed training.
From a cost-benefit perspective, this choice can reduce the delay between data acquisition and method improvement.

Let $C_{\text{delay}}$ denote the additional opportunity cost caused by postponing model update, such as repeated failure or inefficient execution on near-future similar tasks before the model is updated.
Then the cost of a purely delayed training strategy can be written as
\begin{equation}
C_{\text{delayed}} = C_{\text{retrieve}} + C_{\text{exec}} + C_{\text{plan}} + C_{\text{collect}} + C_{\text{train}} + C_{\text{delay}}.
\end{equation}

By contrast, quasi-real-time training attempts to reduce this delay-related term:
\begin{equation}
C_{\text{quasi-real-time}} =
C_{\text{retrieve}} + C_{\text{exec}} + C_{\text{plan}} + C_{\text{collect}} + C_{\text{train}} + C'_{\text{delay}},
\end{equation}
where
\begin{equation}
C'_{\text{delay}} \leq C_{\text{delay}}.
\end{equation}

Hence, although quasi-real-time training may not reduce the direct training cost itself, it can reduce the downstream cost caused by waiting too long before updating the task-solving model.

\section{Verification}
\label{sec:verification}

This section verifies whether the proposed framework can reduce execution time and dependence on the LLM while maintaining task-solving ability.
The verification is conducted from two complementary perspectives.
First, we evaluate repeated-task learning through self-execution, where the robot encounters previously uncovered tasks, learns them, and reuses the resulting local methods in later executions.
Second, we evaluate observation-driven learning, where the robot does not execute the task in the first round but instead observes a successful external behavior and then attempts to transform the observed result into reusable local knowledge.
The purpose of these experiments is to determine whether the proposed framework can convert both execution-derived and observation-derived experience into local reusable capability, thereby reducing repeated dependence on external LLM interaction.

\subsection{Experimental Setup}
\label{subsec:exp_setup}

We constructed a repeated-task action-sequence benchmark.
Each task consists of a natural-language instruction and a corresponding target action sequence selected from the robot's executable action set.
For the self-execution experiment, a total of 20 tasks were generated, and each task was repeated 5 times, resulting in 100 runs for each method.
For the observation-driven experiment, the same repeated-task structure was used, except that the robot did not execute the task in the first round and instead received a successful external observation.

In the self-execution setting, we compared the following three methods:
\begin{itemize}
    \item \textbf{Always-LLM}: the robot queries the LLM for every task execution and does not consolidate a reusable local method;
    \item \textbf{Library-Only}: the robot only searches the local method library and does not trigger new learning when no method is found;
    \item \textbf{Proposed}: the robot first retrieves the local method library, then triggers LLM-driven learning when the task is not covered, and writes the validated result into the local method library for later reuse.
\end{itemize}

In the observation-driven setting, we compared the following two methods:
\begin{itemize}
    \item \textbf{Observation-Only}: the robot observes a successful external behavior in the first round but does not convert the observed result into a reusable local method;
    \item \textbf{Proposed-Observation}: the robot uses the same initial observation, but proactively organizes and consolidates the observed successful behavior into a reusable local method for later reuse.
\end{itemize}

The following metrics are used throughout the verification:
\begin{itemize}
    \item \textbf{Average total execution time}, including planning, LLM interaction, execution, and learning-related overhead;
    \item \textbf{Average LLM calls per task (or per round)}, used to measure dependence on the LLM;
    \item \textbf{Average LLM time ratio}, defined as the fraction of total time spent in LLM interaction;
    \item \textbf{Success rate};
    \item \textbf{Method-library hit rate}, indicating how often a task can be solved directly from the local method library.
\end{itemize}

\subsection{Self-Execution-Based Repeated-Task Verification}
\label{subsec:self_execution_verification}

\subsubsection{Overall Comparison}
\label{subsubsec:self_overall}

\begin{figure}[t]
    \centering
    \includegraphics[width=\columnwidth]{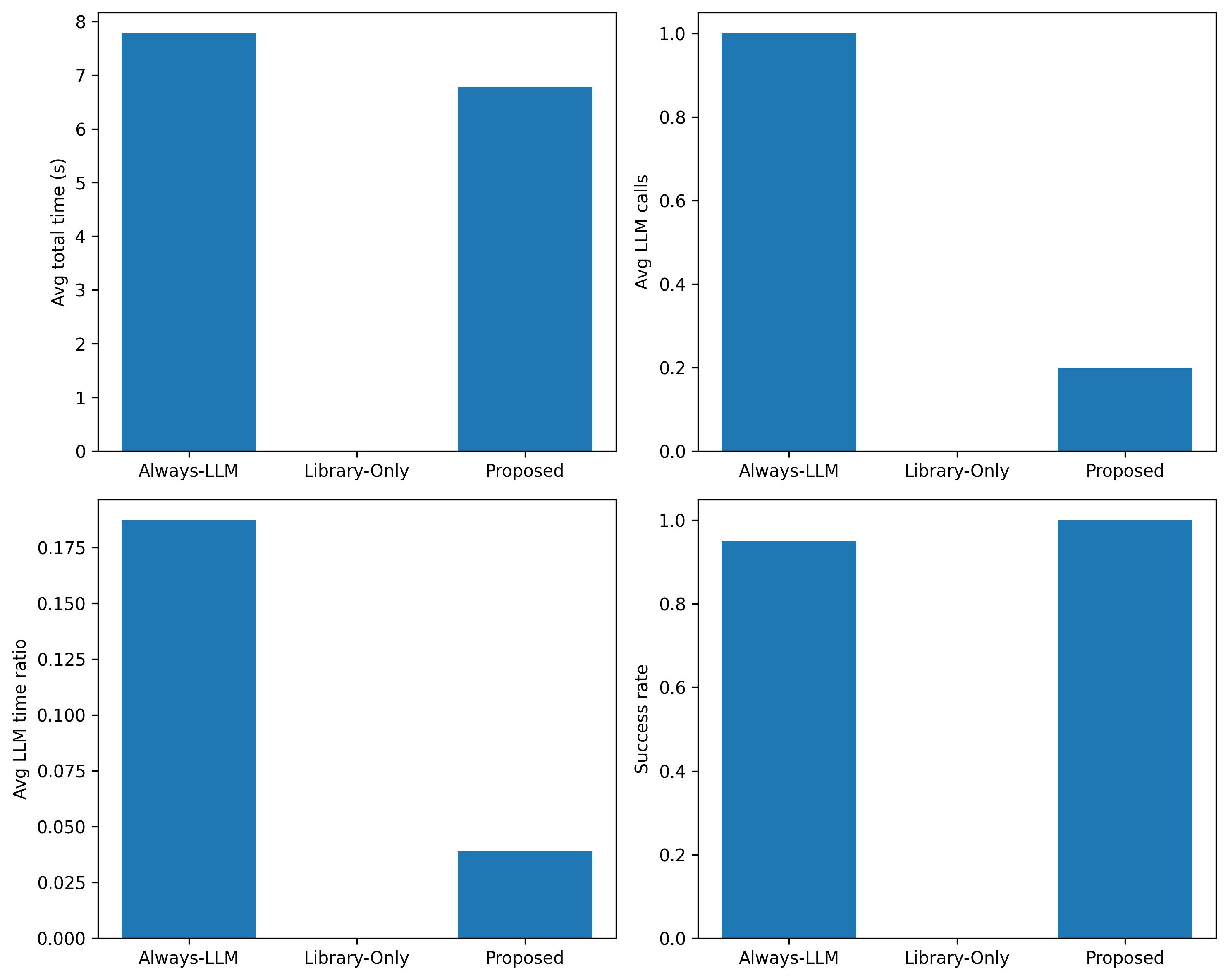}
    \caption{Overall comparison of the three methods in the self-execution repeated-task setting, including average total execution time, average LLM calls, average LLM time ratio, and success rate. The proposed method achieves lower time cost and substantially lower LLM dependence than Always-LLM while maintaining the highest success rate.}
    \label{fig:overall_metrics}
\end{figure}

Fig.~\ref{fig:overall_metrics} shows the overall comparison among the three methods in the self-execution setting.
The proposed framework achieved the best overall balance between efficiency and task completion.
Specifically, the proposed method reached a success rate of $1.00$, compared with $0.95$ for Always-LLM, while the Library-Only baseline failed on uncovered tasks and obtained a success rate of $0.00$.

In terms of efficiency, the proposed method reduced the average total execution time from $7.7772$~s in Always-LLM to $6.7779$~s, corresponding to an improvement of approximately $12.85\%$.
At the same time, it reduced the average number of LLM calls per task from $1.0$ to $0.2$, that is, an $80\%$ reduction.
Similarly, the average LLM time ratio decreased from $0.1873$ to $0.0390$, which is a reduction of about $79.2\%$.
These results indicate that the proposed framework substantially reduces dependence on the LLM after knowledge has been accumulated locally.

It should be noted that the Library-Only baseline has nearly zero execution time and zero LLM calls only because it does not solve uncovered tasks at all.
Therefore, the main meaningful efficiency comparison is between the proposed method and Always-LLM.
Under this comparison, the proposed method shows lower time cost, lower LLM dependence, and slightly higher task success.

\subsubsection{Effect of Repeated Execution}
\label{subsubsec:self_repeat}

\begin{figure}[t]
    \centering
    \includegraphics[width=\columnwidth]{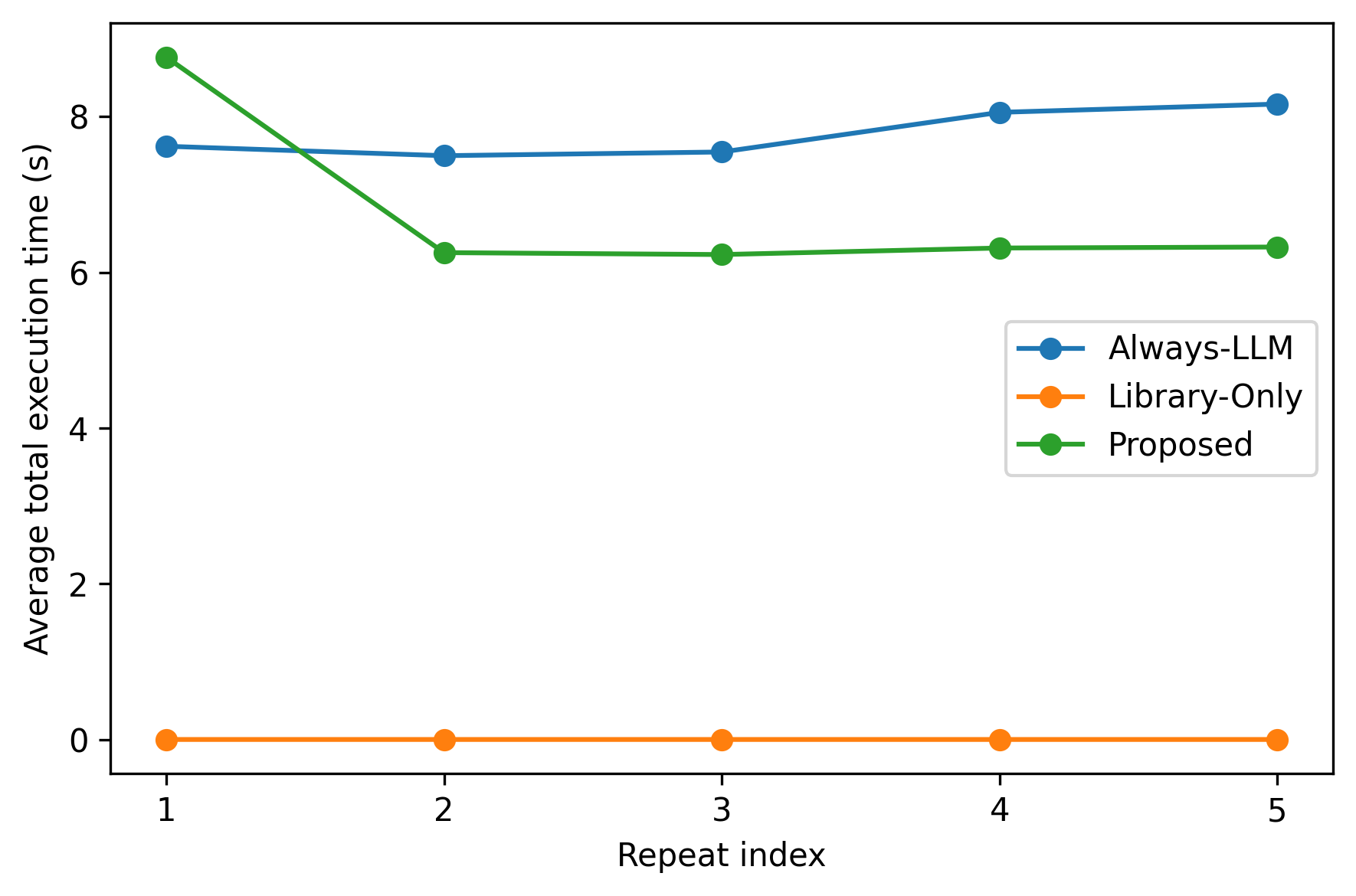}
    \caption{Average total execution time versus repeat index in the self-execution repeated-task setting. The proposed method is initially slower due to learning overhead, but becomes faster than Always-LLM from the second execution onward.}
    \label{fig:repeat_total_time}
\end{figure}

\begin{figure}[t]
    \centering
    \includegraphics[width=\columnwidth]{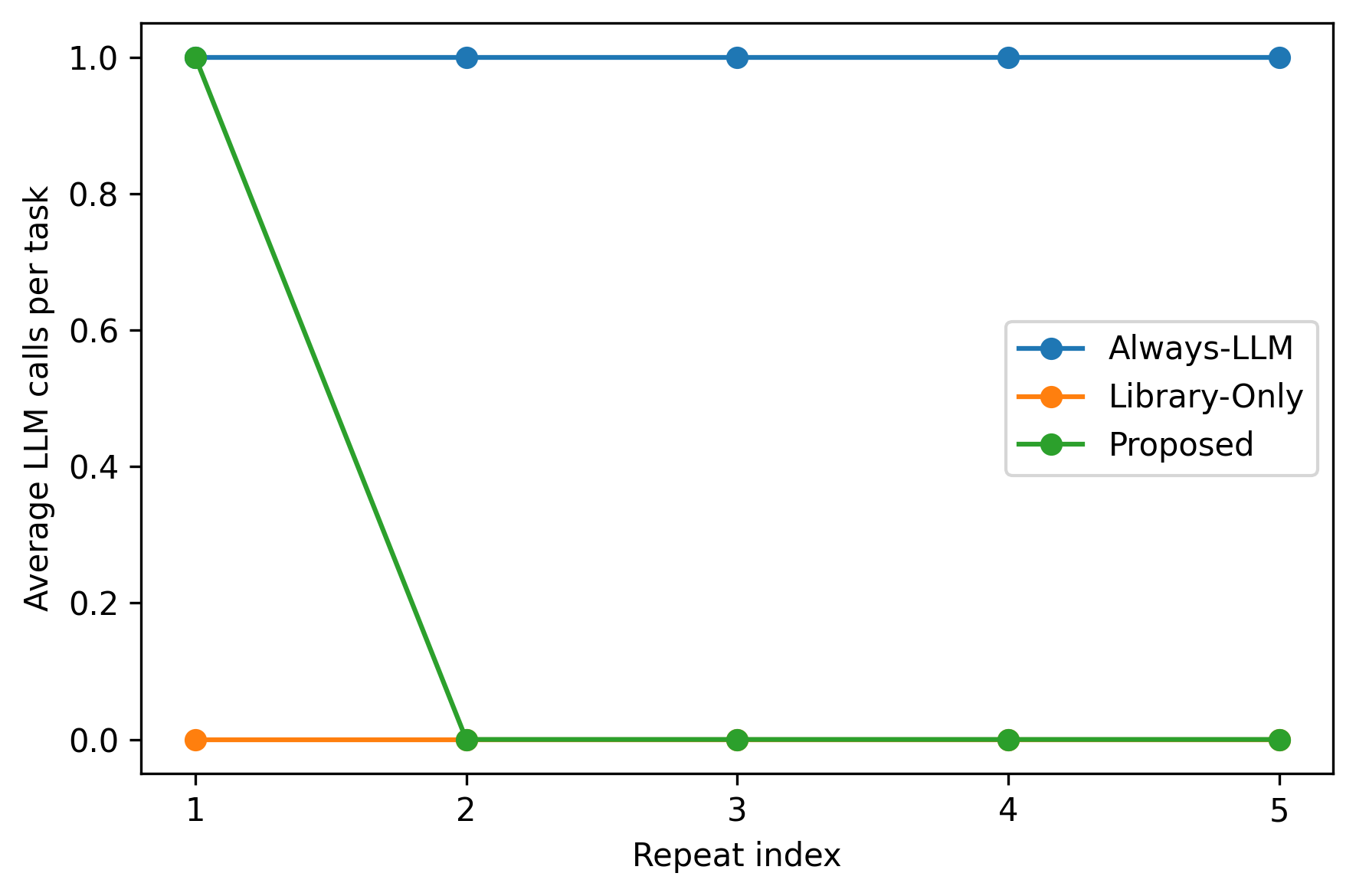}
    \caption{Average LLM calls versus repeat index in the self-execution repeated-task setting. The proposed method requires LLM interaction only in the first execution, whereas Always-LLM depends on the LLM in every execution.}
    \label{fig:repeat_llm_calls}
\end{figure}

\begin{figure}[t]
    \centering
    \includegraphics[width=\columnwidth]{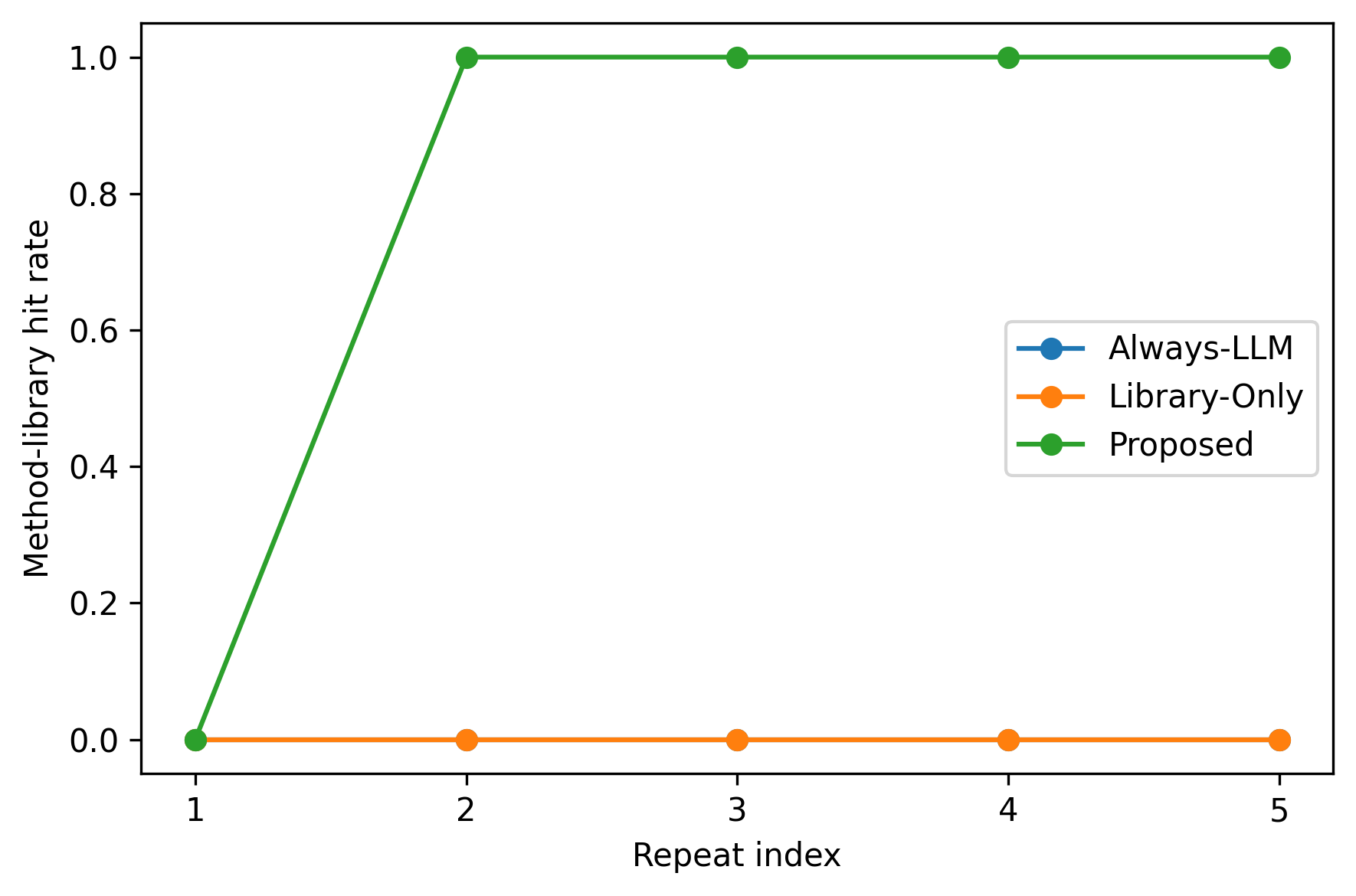}
    \caption{Method-library hit rate versus repeat index in the self-execution repeated-task setting. The proposed method stores learned solutions locally after the first execution, leading to direct reuse from the second execution onward.}
    \label{fig:repeat_hit_rate}
\end{figure}

To verify whether the proposed method can convert an uncovered task into reusable local knowledge, we further analyzed the repeat-wise performance.
Fig.~\ref{fig:repeat_total_time} shows the average total execution time across repeated runs, Fig.~\ref{fig:repeat_llm_calls} shows the average number of LLM calls, and Fig.~\ref{fig:repeat_hit_rate} shows the method-library hit rate.

As shown in Fig.~\ref{fig:repeat_total_time}, the proposed method is slower in the first repeat ($8.7660$~s) than Always-LLM ($7.6208$~s), because it must perform LLM-driven planning and learning for previously uncovered tasks.
However, after the learned solution is stored locally, the proposed method becomes consistently faster, with repeat-2 to repeat-5 average times of $6.2542$~s, $6.2294$~s, $6.3136$~s, and $6.3263$~s, respectively, whereas Always-LLM remains around $7.50$--$8.16$~s.

Fig.~\ref{fig:repeat_llm_calls} further shows that the proposed framework requires one LLM call only in the first repeat and zero LLM calls from the second repeat onward, while Always-LLM depends on one LLM call in every repeat.
This demonstrates that the proposed method effectively reduces repeated reliance on the external LLM after learning.

Moreover, Fig.~\ref{fig:repeat_hit_rate} shows that the method-library hit rate of the proposed framework rises from $0$ in the first repeat to $1.0$ from the second repeat onward.
This confirms that the newly acquired solution is successfully consolidated into the local method library and can be directly reused in later executions.

Overall, these repeat-wise results show that the proposed framework incurs an initial learning cost but subsequently reduces both execution time and LLM dependence through local method reuse.

\subsection{Verification of Observation-Driven Learning}
\label{subsec:obs_verification}

To further verify the observation-driven capability of the proposed framework, we designed an additional experiment in which the robot did not execute the task by itself in the first round.
Instead, it first observed a successful external action sequence and then faced repeated executions of the same task in later rounds.
The purpose of this experiment was to determine whether successful external observation could be proactively transformed into reusable local knowledge, thereby reducing later execution time and dependence on the LLM.

\begin{figure}[t]
    \centering
    \includegraphics[width=\columnwidth]{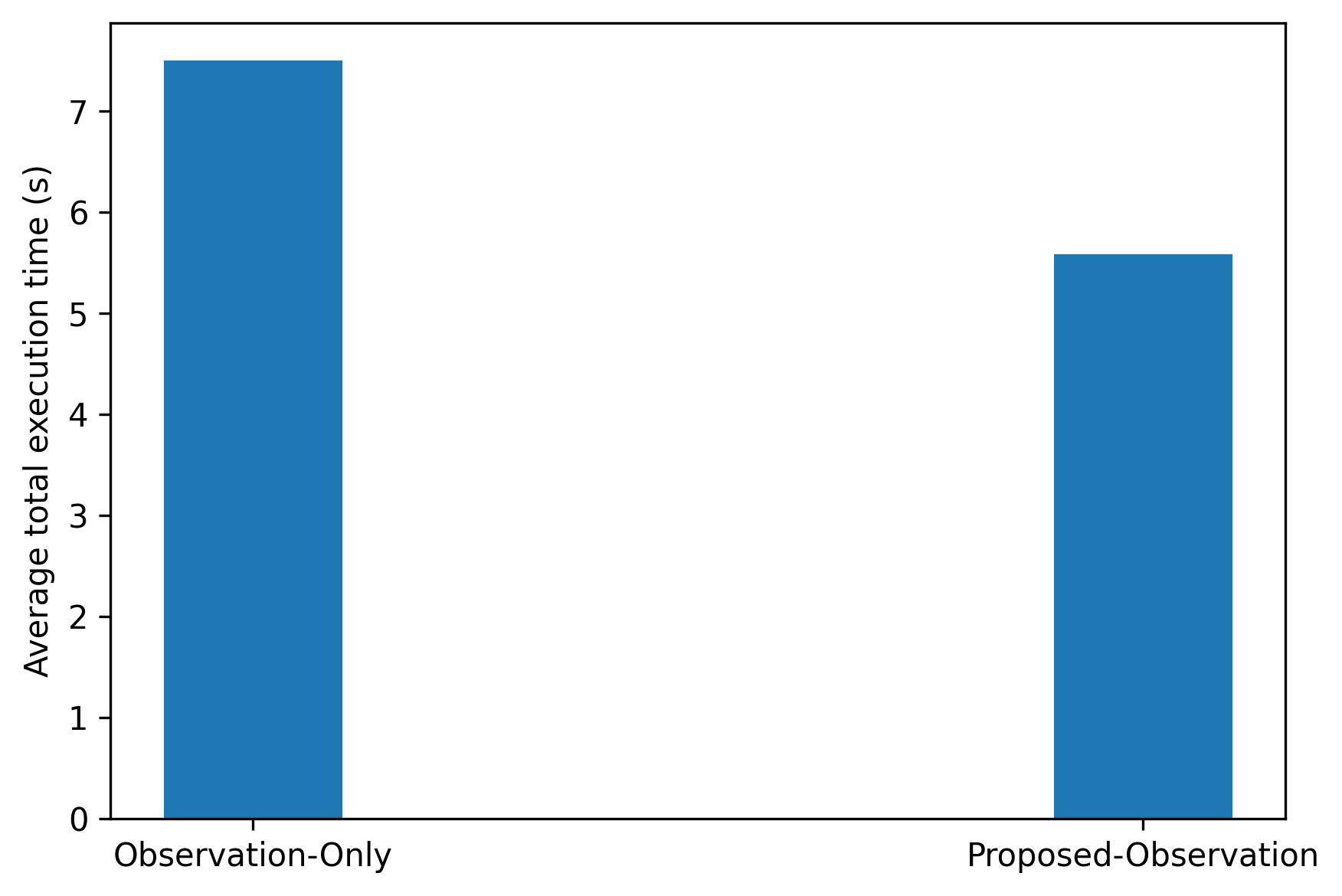}
    \caption{Overall comparison of average total execution time in the observation-driven setting. The proposed observation-driven method is faster than Observation-Only while maintaining the same success rate.}
    \label{fig:obs_overall_total_time}
\end{figure}

Fig.~\ref{fig:obs_overall_total_time} shows the overall comparison of average total execution time in the observation-driven setting.
Both methods achieved a success rate of $1.00$, indicating that the proposed observation-driven mechanism did not reduce task correctness.
However, the proposed method achieved substantially higher efficiency.
Specifically, the average total execution time was reduced from $7.4969$~s in Observation-Only to $5.5833$~s in Proposed-Observation, corresponding to an improvement of about $25.5\%$.
At the same time, the average number of LLM calls per round decreased from $0.8$ to $0.2$, and the average LLM time ratio decreased from $0.2758$ to $0.1866$.
Moreover, the method-library hit rate increased from $0.0$ to $0.8$, showing that the proposed framework successfully converted observed external success into locally reusable knowledge.

\begin{figure}[t]
    \centering
    \includegraphics[width=\columnwidth]{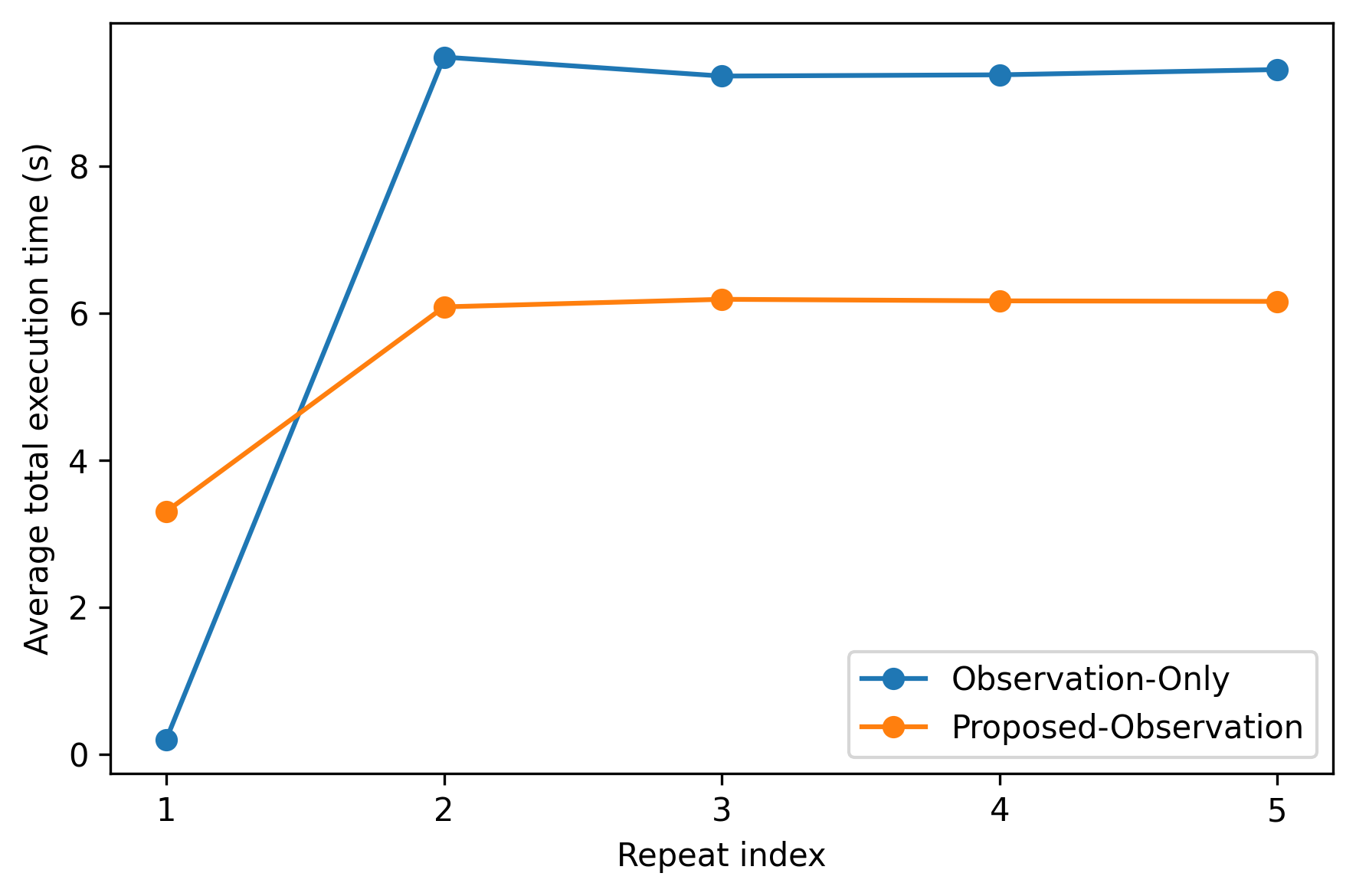}
    \caption{Average total execution time versus repeat index in the observation-driven setting. After the initial observation-and-learning round, the proposed method remains consistently faster than Observation-Only in later executions.}
    \label{fig:obs_repeat_time}
\end{figure}

\begin{figure}[t]
    \centering
    \includegraphics[width=\columnwidth]{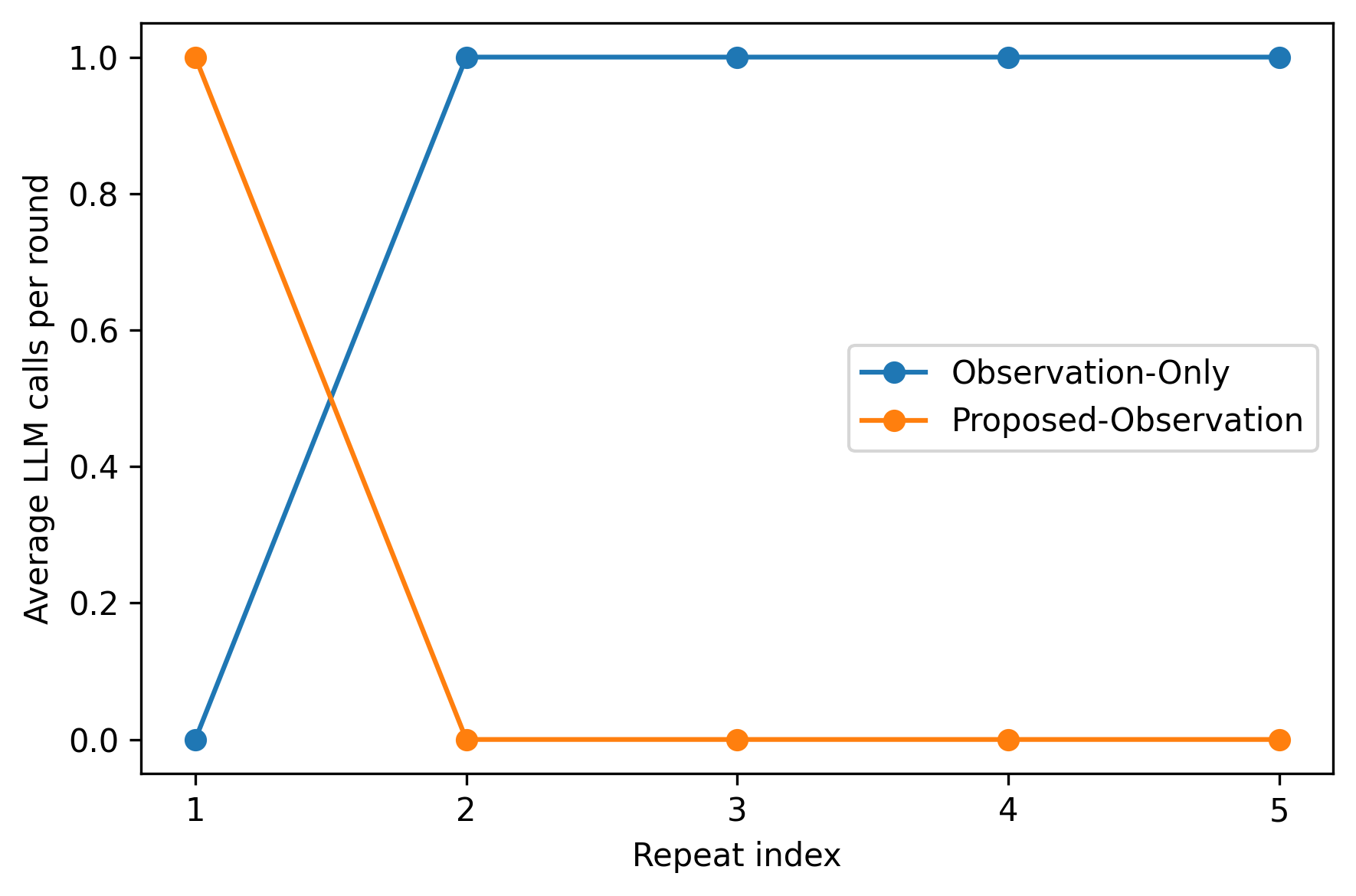}
    \caption{Average LLM calls versus repeat index in the observation-driven setting. The proposed method requires LLM interaction only in the first round, whereas Observation-Only continues to depend on the LLM in later rounds.}
    \label{fig:obs_repeat_llm}
\end{figure}

\begin{figure}[t]
    \centering
    \includegraphics[width=\columnwidth]{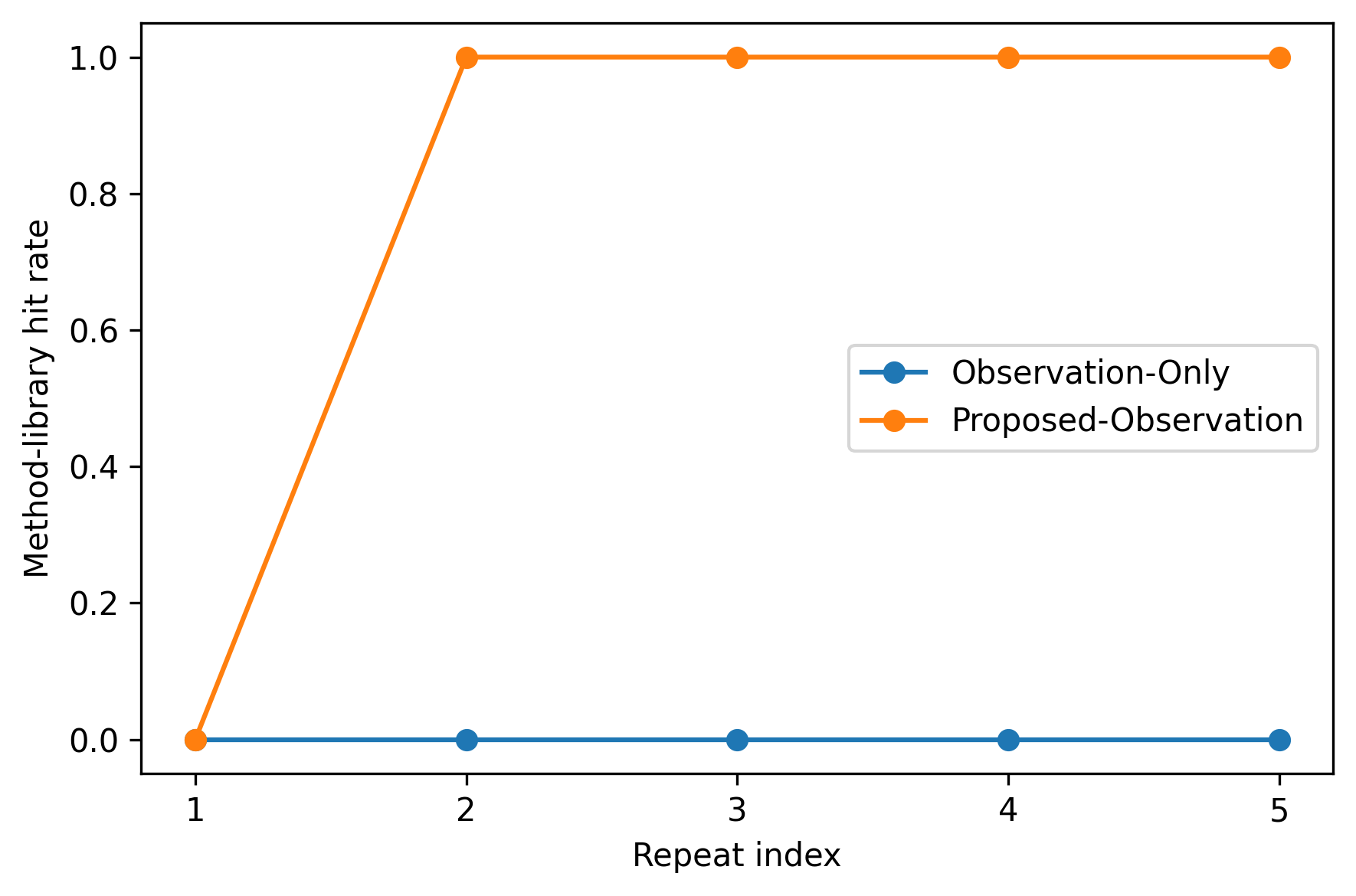}
    \caption{Method-library hit rate versus repeat index in the observation-driven setting. The proposed method successfully converts observed successful behavior into reusable local methods from the second round onward.}
    \label{fig:obs_repeat_hit}
\end{figure}

To better understand the source of this improvement, we further analyzed the repeat-wise behavior, as shown in Fig.~\ref{fig:obs_repeat_time}, Fig.~\ref{fig:obs_repeat_llm}, and Fig.~\ref{fig:obs_repeat_hit}.
For Observation-Only, the first round was only an observation round and thus took only $0.2000$~s, but from the second to the fifth rounds the average total execution time remained high, at $9.4885$~s, $9.2303$~s, $9.2471$~s, and $9.3185$~s, respectively.
In these later rounds, the method still required $1.0$ LLM call per round and its method-library hit rate remained $0.0$ throughout.
This indicates that merely observing successful external behavior is not sufficient if the observed result is not actively transformed into local capability.

By contrast, Proposed-Observation used the first round not only to observe but also to organize and consolidate learning.
As a result, the first round took $3.3031$~s, which was higher than the pure observation cost in Observation-Only because the framework additionally performed LLM-assisted organization and local method construction.
However, from the second round onward, the average total execution time became stable at $6.0884$~s, $6.1915$~s, $6.1705$~s, and $6.1631$~s, respectively, all substantially lower than those of Observation-Only.
Meanwhile, the average number of LLM calls dropped from $1.0$ in the first round to $0.0$ in rounds 2--5, and the method-library hit rate rose from $0.0$ to $1.0$ from the second round onward.
These results show that, once the observed successful behavior is converted into a local reusable method, repeated external LLM interaction is no longer necessary for the same task.

This experiment verifies the observation-driven learning property of the proposed framework.
Even when the robot does not initially execute the task by itself, successful external observation can still be proactively converted into reusable local knowledge.
This conversion enables the robot to reduce both later execution time and dependence on the LLM, while maintaining task success.
Therefore, the proposed framework does not treat observation as a temporary reference only, but as an effective source of autonomous capability acquisition in open environments.

\section{Conclusion}
\label{sec:conclusion}

This paper proposed an LLM-driven closed-loop autonomous learning framework for robots facing uncovered tasks in open environments.
The framework enables the robot to determine when learning should be triggered, how learning should be organized, and how newly acquired experience should be consolidated into reusable local knowledge.
Unlike approaches that repeatedly depend on external LLM interaction, the proposed method supports capability accumulation through both self-execution and active observation.

The verification results showed that the proposed framework effectively reduced execution time and dependence on the LLM while maintaining task success.
In the repeated-task self-execution setting, it reduced the average total execution time from 7.7772\,s to 6.7779\,s and the average number of LLM calls per task from 1.0 to 0.2.
In the observation-driven setting, it further reduced the average total execution time from 7.4969\,s to 5.5833\,s and the average number of LLM calls from 0.8 to 0.2.

Overall, the proposed framework provides a practical way for robots to gradually transform both execution-derived and observation-derived experience into reusable local capability, thereby reducing repeated external dependence in long-term open-environment operation.


\ifCLASSOPTIONcaptionsoff
  \newpage
\fi

\bibliographystyle{IEEEtran}
\bibliography{ref}

%

\begin{IEEEbiography}{Hong Su}
  received the MS and PhD degrees, in 2006 and 2022, respectively, from Sichuan University, Chengdu, China. He is currently a researcher of Chengdu University of Information Technology Chengdu, China. His research interests include blockchain, large language model and human simulation computing.
\end{IEEEbiography}




\end{document}